\documentclass[a4paper, 10pt, conference]{ieeeconf}      

\IEEEoverridecommandlockouts                              

\usepackage{graphics} 
\usepackage{graphicx}
\usepackage{amsmath} 
\usepackage{comment}
\usepackage{subcaption} 
\usepackage{subcaption} 
\usepackage[compatibility=false]{caption} 
\usepackage{booktabs} 
\usepackage{tabularx} 
\usepackage{array} 
\usepackage{lipsum}
\makeatletter
\let\NAT@parse\undefined
\makeatother

\usepackage[hidelinks]{hyperref} 

\title{\LARGE \bf
Closing the Lab-to-Store Gap: A Data-Efficient Post-Training and Experience-Driven Learning VLA Framework for Retail Humanoids}

\author{Roger~Sala~Sisó$^{1,2,*}$, Tiago~Silvério
$^{1,*}$, Jakob~Sand$^{1}$, Tran~Nguyen~Le$^{2}$ \\[.5em]
\footnotesize\normalfont
    $^{*}$ Equal Contribution \quad
    $^{1}$ HIVE Robots, Denmark \quad
    $^{2}$ Department of Engineering Technology, Technical University of Denmark, Denmark}
\begin{document}

\maketitle
\thispagestyle{empty}
\pagestyle{empty}

\begin{abstract}

Closing the gap between benchmark performance and reliable real-world operation remains a central challenge for Vision-Language-Action (VLA) humanoid robots, which must handle execution errors, distribution shifts, and environmental variability. This paper presents DEED (Data-Efficient Post-Training and Experience-Driven Learning), a systems-level approach evaluated on a supermarket chip-restocking task using a Unitree G1-Edu humanoid robot and the GR00T N1.6 foundation model. DEED comprises three key components: (1) a data-efficient post-training pipeline with control-frequency alignment, data curation, task-relevant visual highlighting, and reduced VLA dependence; (2) a real-world study of experience-driven refinement, adapted from RECAP via a text-based advantage prefix and a vision-language value function; and (3) a latent-space analysis tool for studying in- and out-of-distribution behavior. Our results suggest that bridging the lab-to-store gap is primarily a systems integration challenge rather than an architectural one: careful data design and targeted post-training can transform a policy that fails under naive fine-tuning into a competent real-world system using only a single GPU.

\end{abstract}


\section{Introduction}
\label{sec:introduction}
Vision-Language-Action (VLA) models have emerged as the dominant
paradigm for general-purpose robotic manipulation by extending
internet-scale vision-language foundation models with action
prediction~\cite{kim2024openvla,bjorck2025gr00t}. Models such as
OpenVLA, the $\pi$ family, and GR00T demonstrate strong generalization across objects, instructions, environments, and humanoid embodiments~\cite{kim2024openvla,black2024pi0,bjorck2025gr00t}. As these models become increasingly capable across embodiments, attention has shifted toward humanoid robots, whose human-like morphology enables them to leverage existing infrastructure and perform manipulation tasks in environments designed for people, such as retail stores.

\begin{figure}[t]
    \centering
    \includegraphics[width=1\linewidth, trim=0 0 0 20px, clip]{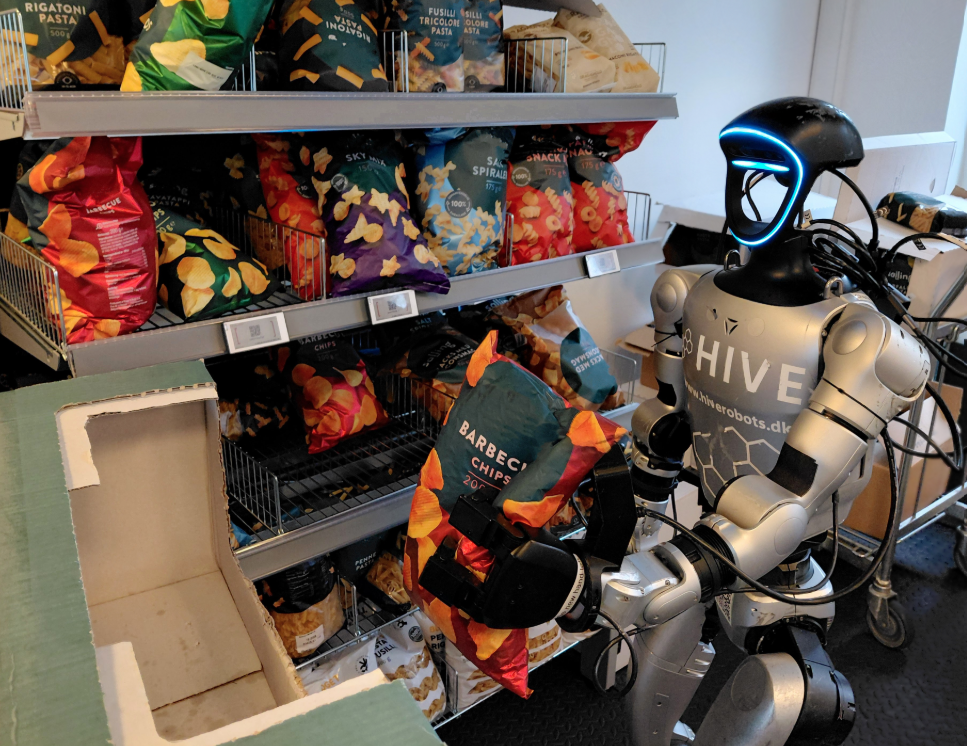}
    \caption{Hardware setup used for data collection and the experiments. The system consists of a Unitree G1-Edu equipped with two external wrist-mounted cameras and is evaluated on a supermarket restocking task.}
    \label{fig:robot}
    \vspace{-.6cm}
\end{figure}

Despite this progress, a substantial gap remains between benchmark performance and reliable deployment. Most VLAs are trained offline and deployed as fixed policies, whereas practical deployment demands robustness to execution errors, distribution shift, and continual environmental variability \cite{kim2024openvla}. Two challenges dominate. First, deploying a pretrained VLA depends less on architectural novelty than on engineering choices such as action representation, camera placement, control-frequency alignment, and data curation; since behavioral cloning is fundamentally limited by its demonstrations, data inconsistencies or modest distribution shift quickly accumulate into policy failures \cite{belkhale2023data}. Second, offline policies cannot improve from their own experience, motivating post-training methods for continual adaptation after deployment \cite{physical2025pi0.6}. 

Recent work addresses these challenges from complementary directions. Data-centric studies show that action diversity, transition diversity, and dataset quality strongly influence imitation performance \cite{belkhale2023data}, while visual highlighting, long-horizon task decomposition, error recovery, and improved evaluation protocols target common failure modes \cite{hannus2025iavla,hdspace2025,rac2025,wang2025shortcut,zhang2025experiences}. Interaction-based methods refine pretrained VLAs through online reinforcement learning; in particular, RECAP integrates RL signals into the imitation objective via advantage conditioning, avoiding unstable policy-gradient updates on large pretrained models \cite{physical2025pi0.6}. Complementary in/out-of-distribution detection methods provide a mechanism for monitoring deployment reliability \cite{lee2018simple}.
 
This paper adopts a systems perspective on deploying VLA-based humanoids, motivated by an ongoing industrial effort to deploy
humanoid restocking in an operating supermarket, whose practical
constraints shape our design. We present \textbf{DEED} (Data-Efficient Post-Training and Experience-Driven Learning), a framework for robust real-world deployment, instantiated on a supermarket chip-restocking task with a Unitree G1-Edu and the GR00T N1.6 foundation model. DEED has three components. First, a data-efficient post-training pipeline capturing the practical engineering needed to convert an out-of-the-box GR00T checkpoint into a reliable policy: frequency alignment, data curation, task-relevant visual highlighting, and techniques for reducing VLA dependence. Second, an empirical study of RECAP-style refinement on a real humanoid, adapted to GR00T's decoupled architecture through a text-based advantage prefix and a vision-language only value function. Third, a latent-space in/out-of-distribution analysis method measuring how far deployment states deviate from the policy's training distribution.

Finally, we provide an empirical study of experience-driven post-training in a real retail setting, identifying where self-generated experience improves robustness and where repeated refinement begins to degrade performance as self-generated rollouts dominate the training distribution, yielding practical guidance for deploying foundation-model-based humanoids in the real world.
    



\section{DEED Framework}
\label{sec:method}
\subsection{Data-Efficient recipe for VLA Post-Training}
\label{sec:DE_recipe}

Finetuning a VLA for real-world deployment involves more than collecting demonstrations and training a model. This section presents the Data-Efficient (DE-) post-training recipe of DEED: the practical decisions and tricks we found necessary when finetuning a VLA (here GR00T N1.6) out of the box, which we believe help anyone in this position avoid common pitfalls.

\paragraph{\textbf{Frequency Hierarchy}}
One of the most impactful early decisions is establishing a clear hierarchy between camera frequency ($f_{cam}$), teleoperation frequency ($f_t$), recording frequency ($f_r$), and inference control frequency ($f_{ctrl}$), which must satisfy
\begin{equation}
    f_r = f_{\mathrm{ctrl}} \leq f_{\mathrm{cam}},
    \qquad
    f_t \geq f_r.
\end{equation}
Vision is the policy's only channel for perceiving the world, so recording or acting faster than the camera means learning from duplicated observations that have not changed between frames, producing behavior that is neither smooth nor reliable at inference. Recording frequency in turn determines the spatial resolution between consecutive action targets, which dictates the smoothness of the motion the policy learns, and higher frequency allows continuous movements with stable velocities, crucial for manipulation tasks~\cite{wang2026learning}. Recording and inference control frequencies must match, since the policy learns the temporal semantics of the action space at whatever rate it was recorded at, and dispatching actions at any other rate at inference time is a mismatch the model was never trained to handle. Teleoperation frequency does not need to respect the camera ceiling: in slowly evolving scenes $f_t$ can exceed $f_{cam}$ without meaningful visual information being lost, and keeping $f_t$ high improves the operator's embodiment and teleoperation performance, bounded only by the teleoperation system's own compute and latency budget.

\paragraph{\textbf{Data Curation Rules}}
Imitation learning is a direct observation-to-action mapping with no implicit notion of intent or of which aspects of the observation are task-relevant, so every demonstration is learned with equal weight. Data curation is therefore crucial to prevent noise and systemic bad patterns from compromising learning. Each rule below corresponds to a failure mode we encountered and mitigated during the development of the training pipeline.

\begin{enumerate}
  \item \textbf{Ensure balanced state coverage.} Unbalanced datasets bias the policy toward the actions and dynamics of dominant states, degrading performance in underrepresented ones even when included in training.
  \item \textbf{Record only efficient, successful behavior.} Avoid hesitations, corrections, backtracking, or inefficient paths, and trim episodes at the point where failure becomes evident rather than recording the lead-up to a failed attempt as valid behavior~\cite{kim2024openvla}.
  \item \textbf{Retain incidental recoveries.} A policy trained purely on clean successes has no learned behavior for returning to the task once prediction errors compound and drift it into poorly represented states~\cite{belkhale2023data}; we trim only the failed segment of an episode and keep the recovery, letting the policy learn how to respond when it departs from normal execution.
  \item \textbf{Maximize action consistency, not just state coverage.} If the expert's actions vary inconsistently across similar states, the policy lacks a stable target and either converges to an averaged, ineffective action or learns an unintended observation-action mapping~\cite{belkhale2023data}; prioritize consistent actions at repeated states over covering many states once each.
  \item \textbf{Minimize uncontrolled environmental variation.} When observation changes are driven by external factors (objects shifting, lighting fluctuations, other agents), recorded transitions no longer reflect a consistent action-outcome relationship~\cite{belkhale2023data}; stabilize the physical setup during collection rather than expecting training to filter this out.
  \item \textbf{Avoid no-operation states.} Idle segments at episode boundaries pair effectively unchanged states with inconsistent actions (no-op during idling, then a real action once motion begins) and encourage less dynamic behavior~\cite{kim2024openvla}; start episodes once meaningful motion begins and end them as soon as the goal is achieved.
\end{enumerate}

\begin{figure}[t]
    \centering
    \begin{subfigure}{0.48\linewidth}
        \includegraphics[width=\linewidth, trim=0 0 0 64px, clip]{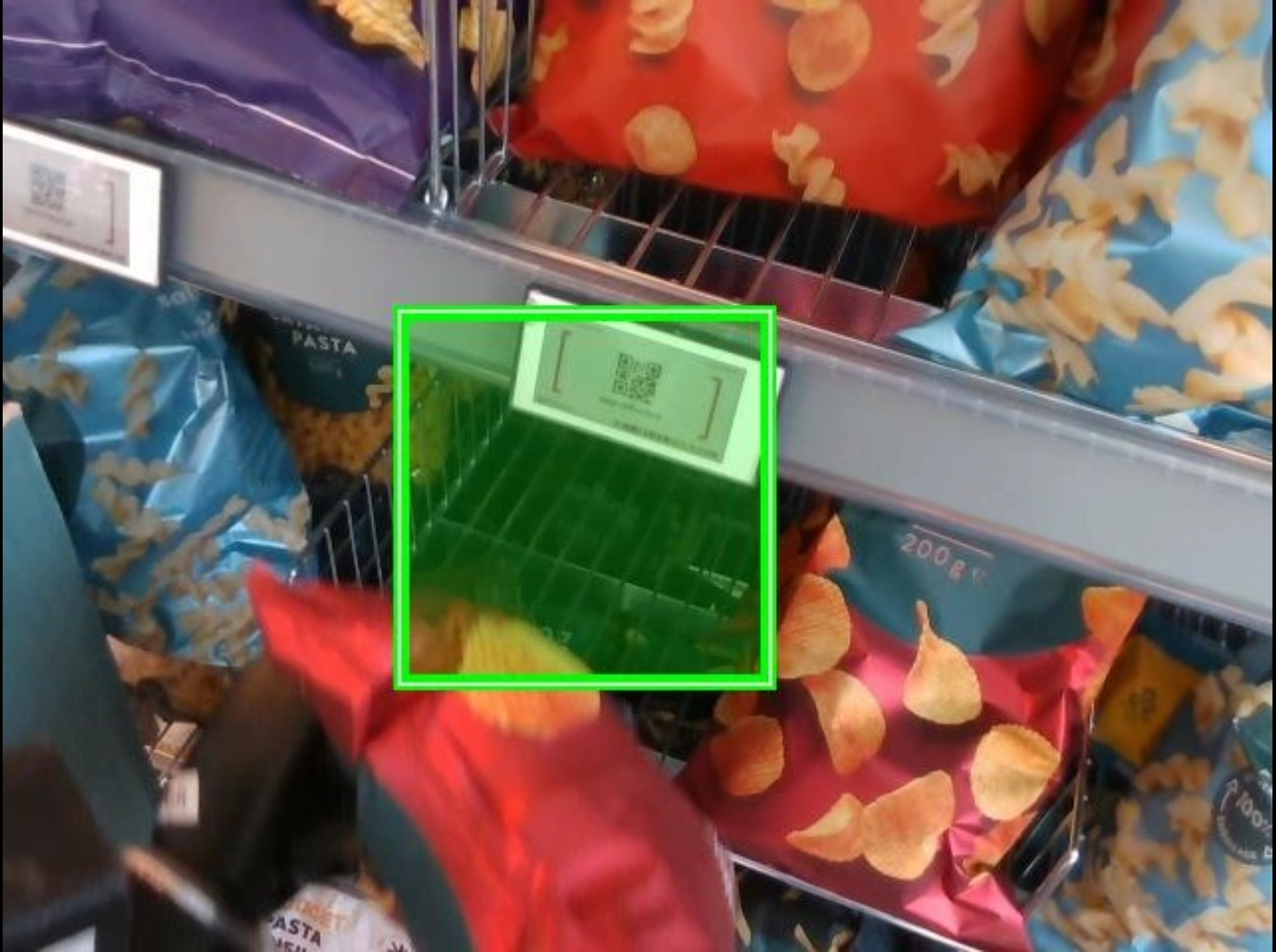}
        \caption{IA-VLA Activated}
    \end{subfigure}
    \begin{subfigure}{0.48\linewidth}
        \includegraphics[width=\linewidth, trim=0 0 0 64px, clip]{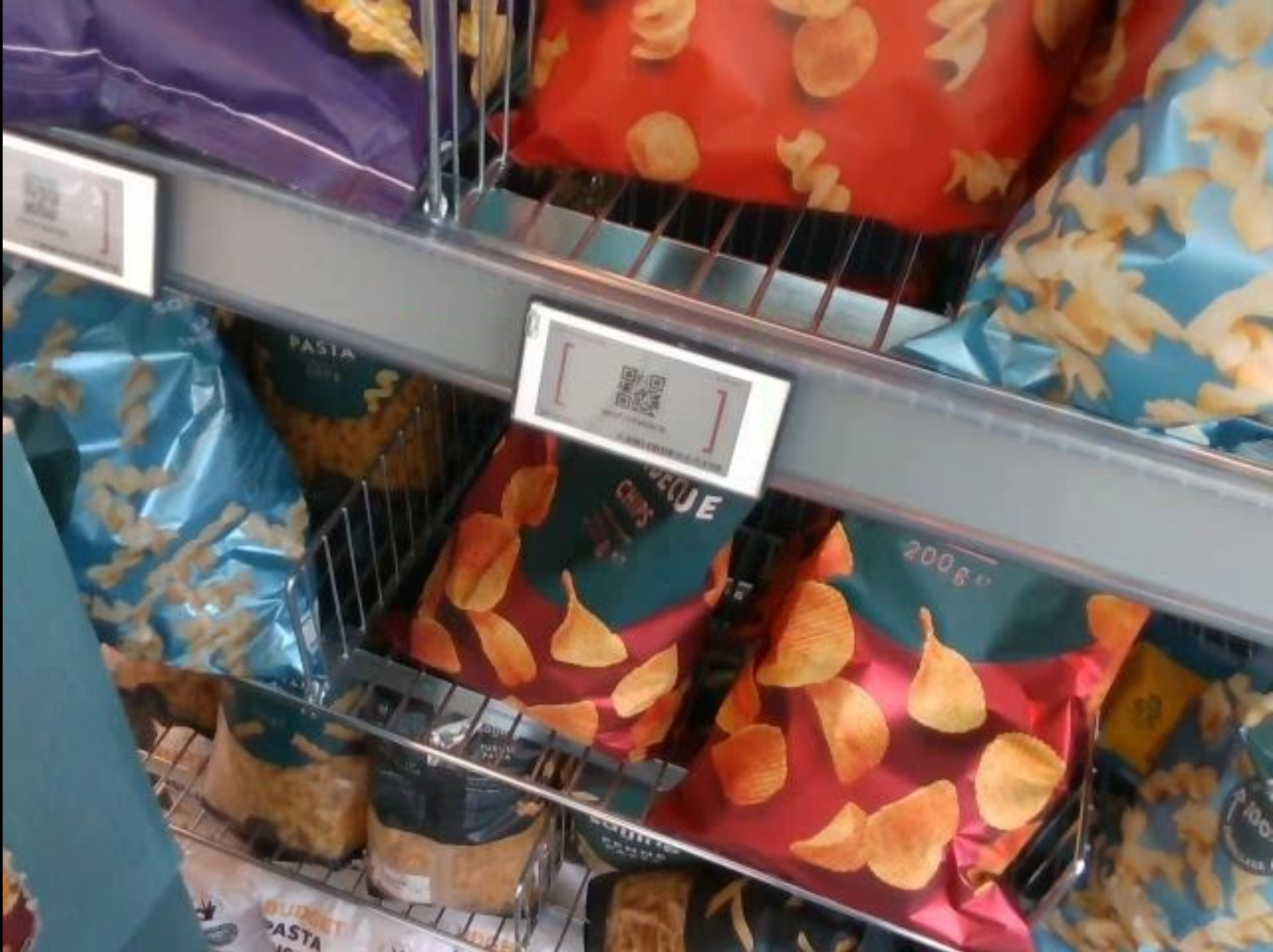}
        \caption{IA-VLA Deactivated}
    \end{subfigure}
\caption{Head camera image processed with IA-VLA during inference on the experiment presented in Section~\ref{sec:experiments}: (a) the placing area is activated as a green bounding box, adapted per frame, highlighting the empty shelf location as the target for placement; (b) the mask is deactivated after the item is successfully placed.}
\label{fig:IA-VLA}
    \vspace{-.6cm}
\end{figure}

\paragraph{\textbf{Highlighting Task-Relevant Visual Cues.}} Vision is the primary modality through which the policy perceives the world, making extraction of task-relevant visual features critical. This is especially challenging for VLAs, whose visual backbones are pretrained on internet-scale rather than embodied, task-specific data, leaving them prone to misleading correlations such as background texture, viewpoint, or task-irrelevant objects \cite{wang2025shortcut}. 
We integrate an adapted IA-VLA framework \cite{hannus2025iavla} into our pipeline (Figure~\ref{fig:IA-VLA}): a large VLM highlights task-relevant image regions via segmentation masks propagated across frames, avoiding per-timestep inference, and giving the VLA task-relevant visual grounding rather than learning it from post-training data alone.
These masks identify task-relevant objects; the placing area is then the empty region from their complement, highlighted with a bounding box.

\paragraph{\textbf{Practical Tricks to Reduce VLA Dependence.}}

As VLAs remain at an early stage of real-world readiness, we suggest two practical tricks that enhance their applicability on humanoids:

\begin{enumerate}
  \item \textbf{Binary Hand Controller.} Manipulation demands high-frequency, fine-grained adaptation to physical interactions. Current foundation models, constrained by low-frequency inference and purely image-based perception, do not yet meet this requirement reliably in end-to-end control \cite{liu2025forcevla,wang2026compliantvla}. For tasks reducing primarily to opening and closing the hand, we recommend controlling the hand as a gripper via a binary signal rather than learning continuous, fine-grained joint control. A binary decision correlates strongly and reliably with visual features, especially with wrist-mounted cameras, avoiding continuous actuation for an inherently discrete action.
  \item \textbf{Action Smoothing via Butterworth Filtering.} Transitions between consecutive action chunks are a common source of jerky movement. When the model overfits, chunk transitions become artificially smooth, but at the cost of reduced generalization at both the action head and visual representation level. Since the training data consists of smooth trajectories, jerkiness pushes executed actions out of the training distribution, producing erratic or unreliable execution. We therefore apply a Butterworth filter to the predicted action sequence at inference time, a simple and effective solution that preserves the policy's generalization while keeping executed actions smooth and in-distribution.
\end{enumerate}

\paragraph{\textbf{Continuous Episodes for Multi-Subtask Demonstrations}} For multi-subtask tasks, subtask transitions are a common failure point, since segmented demonstrations implicitly train the model to expect episode boundaries as points of static, repeated actions (GR00T pads target chunks by repeating the last in-episode action)~\cite{hdspace2025}. Rather than the common strategy of segmenting long-horizon demonstrations with overlapping boundaries, we recommend recording continuous episodes that span full subtask transitions, which also simplifies data collection. We further advise avoiding shared wording across subtask language prompts, as a weak language signal can let the policy continue a previous subtask or execute the next one with degraded quality despite an unchanged prompt.

\paragraph{\textbf{Policy Evaluation.}} A single aggregate success rate gives an incomplete picture of a VLA's capabilities, failure modes, and generalization, especially for humanoids where embodiment dimensionality and whole-body coordination introduce additional distribution shift; meaningful evaluation therefore requires qualitative analysis alongside quantitative metrics. Although we do not report a dedicated evaluation across generalization regimes, the framework of \cite{zhang2025experiences} guided the data-engineering recipe: distinguishing in-distribution (ID), spatial OOD, and instance+spatial OOD scenarios helped identify the variability to incorporate during data collection, encouraging task-level understanding rather than memorization of specific environments.

\subsection{Experience-Driven Learning Strategy}

The second component of DEED is the Experience-Driven (-ED) stage of the training pipeline. It addresses a fundamental limitation of VLAs: a policy obtained through supervised fine-tuning on teleoperated demonstrations can only be as good as the data it was shown and has no mechanism to improve from its own deployment experience. To close this gap, we adapt the RECAP (RL with Experience and Corrections via Advantage-conditioned Policies) recipe, originally introduced with $\pi^{*}_{0.6}$, to the decoupled GR00T architecture used throughout DEED \cite{physical2025pi0.6,bjorck2025gr00t}.

Adapting RECAP to GR00T poses an architectural challenge: unlike the tightly coupled VLA for which RECAP was proposed, GR00T separates perception and reasoning (a Vision-Language backbone) from control (a Flow-Matching action head). We address this by conditioning the policy through a text-based instruction prefix (``Advantage=True/False'') and restricting the value function to camera observations and text instructions. The refinement procedure thus requires no architectural modification to the pretrained model and should be applicable to other VLA architectures.

\paragraph{\textbf{Pipeline overview}}

\begin{figure}
    \centering
    \includegraphics[width=1\linewidth]{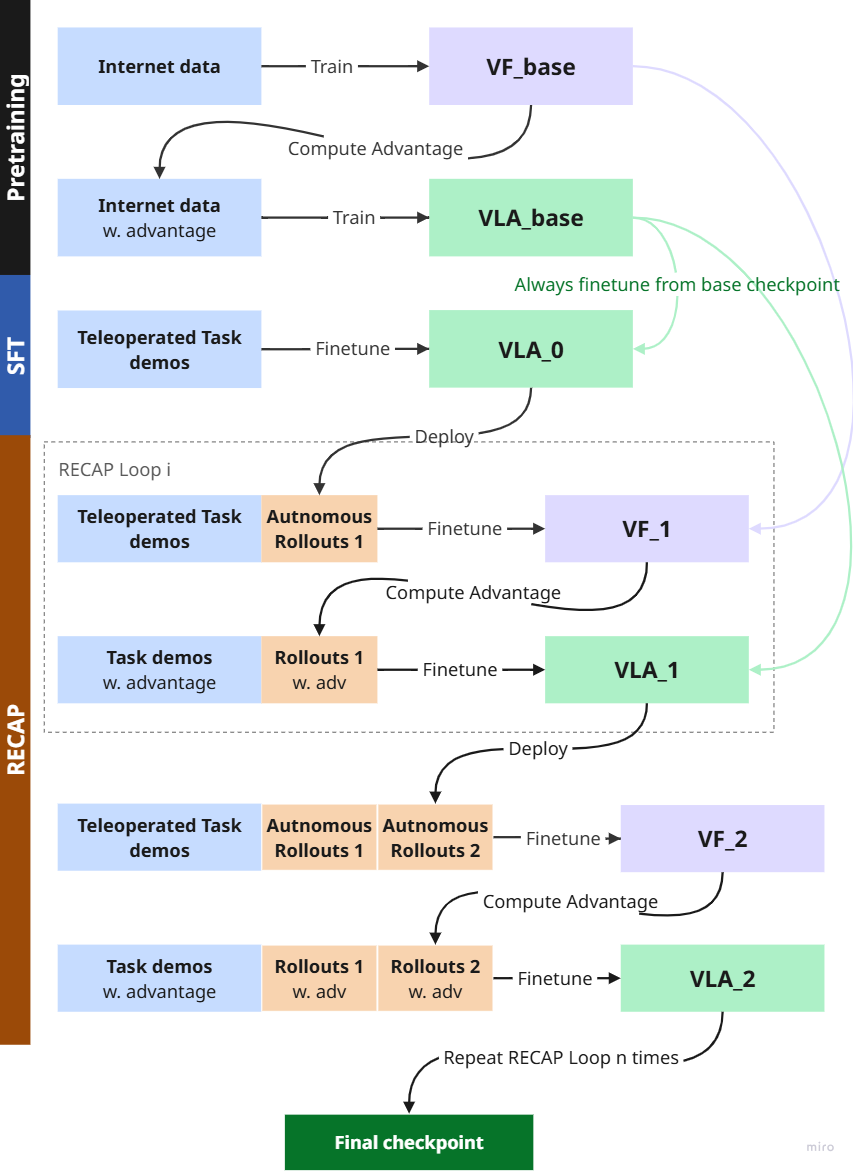}
    \caption{Overview of the Experience-Driven Policy Refinement pipeline. Starting from a policy trained via supervised fine-tuning (SFT) in the data-efficient stage of the DEED framework, RECAP is used to further refine the policy using data collected from autonomous rollouts. At each refinement iteration, both the value function and the VLA model are initialized from their respective pre-trained checkpoints rather than from the previous RECAP stage.}
    \label{fig:recap}
\end{figure}

The Experience-Driven strategy follows an iterative refinement procedure. Starting from the policy obtained with the Data-Efficient recipe, autonomous rollouts are collected on the target task while a human operator intervenes only when necessary to correct erratic or failing behavior. The resulting dataset contains the original teleoperation episodes, successful autonomous executions, and human recovery demonstrations, allowing the policy to learn not only from offline teleoperated data but also from its own interactions with the environment. Figure~\ref{fig:recap} gives an overview of this refinement loop. The pretraining stage shown in the
figure follows the original RECAP formulation and lies
outside the scope of this work: we build directly on the publicly
available pretrained checkpoint rather than reproducing its
computationally intensive pretraining.

Following RECAP, a value function $V_\phi(o_t, \ell)$ is trained on the combined dataset of autonomous rollouts, teleoperation demonstrations, and human corrective interventions to estimate the expected return of each visual observation $o_t$ under instruction $\ell$. The learned values are used to compute an advantage label $\hat{A}_t$ for every timestep, acting as a conditioning token during training. While advantage labels provide outcome-based supervision, human corrective interventions are treated as explicit recovery behaviors and receive a positive supervision signal regardless of estimated advantage, encouraging corrective actions that guide the robot back to successful completion while advantage conditioning suppresses behaviors associated with poor outcomes.
 
To improve optimization stability, each refinement iteration is initialized from the original checkpoint (\texttt{GR00T-N1.6-G1-PnPAppleToPlate}) rather than the previous policy. As observed in \cite{physical2025pi0.6}, restarting from the base checkpoint mitigates optimization drift and prevents error accumulation across refinement cycles, while each iteration still benefits from newly collected experience. During inference, the advantage label is fixed to \texttt{True}, so the deployed policy corresponds to $\pi_\theta(a_t \mid o_t, \ell, A{=}\texttt{True})$, biasing the robot toward actions associated with positive advantage and consistent task progress.

\paragraph{\textbf{Value function}}

 The value function estimates the expected future return of each visual observation encountered during execution. Following RECAP, predicted values are used exclusively to derive advantage labels for training and are never queried at inference, so the value model is optimized independently of the policy without affecting deployment efficiency.
 
To keep it VLA-agnostic, the value function conditions only on the language instruction $\ell$ and visual observation $o_t$, not the robot state. It reuses the pretrained Eagle-3 Vision-Language encoder $f_\psi$ employed by GR00T, followed by a lightweight MLP value head with two 1024-unit hidden layers and GELU activations, applied to the mean-pooled token embeddings. Rather than regressing a scalar directly, the head predicts a categorical distribution $p_\phi(\cdot \mid o_t, \ell)$ over $B = 101$ uniformly discretized return bins with centers $\{b_i\}_{i=1}^{B}$ spanning the normalized interval $[-1, 0]$, with the scalar estimate recovered as the expectation:

\begin{equation}
  V_\phi(o_t, \ell) = \sum_{i=1}^{B} p_\phi(b_i \mid o_t, \ell) \, b_i.
\end{equation}

Supervision uses Monte Carlo returns from complete trajectories, with per-step reward $r_t = 0$ on terminal success, $r_t = -C_{\text{fail}}$ on terminal failure ($C_{\text{fail}}=80$), and $r_t = -1$ otherwise, so the return $R_t = \sum_{k=t}^{T} r_k$ decreases linearly with time-to-go on successful episodes and collapses to a separated lower-value regime on failures. Because episode durations vary with the number of objects manipulated, returns are normalized by the per-task mean trajectory length $s_\ell$ rather than a fixed horizon. The head is trained by cross-entropy against the one-hot target bin $y_i$ of the normalized return:

\begin{equation}
  \mathcal{L}_{\text{value}} = -\sum_{i=1}^{B} y_i \log p_\phi(b_i \mid o_t, \ell).
\end{equation}

The Eagle-3 encoder $f_\psi$ remains frozen throughout, and image embeddings are precomputed and cached, so training updates only the MLP head $g_\phi$, substantially reducing cost while preserving the pretrained visual representations. Training takes approximately 1.5 hours for 5,000 steps at batch size 16 with AdamW, learning rate 1e-4, weight decay 1e-4, and a 15\% held-out validation split.

\paragraph{\textbf{Advantage computation}}

After training the value function, advantage labels are computed offline for every action. Rather than a one-step temporal-difference estimate, we adopt the multi-step estimator from RECAP with lookahead horizon $N=25$,
\begin{equation}
\hat{A}^{(N)}_t =
  \frac{R_t - R_{t+N}}{s_\ell} + V_\phi(o_{t+N}, \ell) - V_\phi(o_t, \ell),
\end{equation}
with the truncated form $R_t/s_\ell - V_\phi(o_t,\ell)$ when $t+N \geq T$. This balances realized task progress over the horizon against the critic's estimate of future improvement, providing a more reliable measure of long-term benefit than instantaneous feedback. Continuous values are converted into binary labels at a threshold $\tau$ chosen so that a target fraction $\epsilon = 0.4$ of the dataset is labeled positive.

As in RECAP, all human corrective interventions are force-labeled positive ($c_t = 1$) regardless of their estimated advantage, ensuring recovery behaviors are consistently reinforced.

Labels are encoded as the text prefixes \texttt{Advantage=True/False} and concatenated to the front of the instruction $\ell$ before flow-matching fine-tuning. A conditioning dropout of $p_{\text{drop}} = 0.3$ is applied during dataset generation, rather than during training, to avoid modifying the GR00T training pipeline. Since the dataset contains conditioned and unconditioned samples, the policy retains its original imitation capabilities while learning to preferentially imitate positively labeled actions. At inference the label is fixed to \texttt{True}, biasing the policy toward behaviors with higher expected task success.

\subsection{In/Out-Distribution Analysis Tool}

The final component of DEED is a tool for measuring, at runtime or offline, how far the robot's current configuration lies from the states seen during data collection. Rather than being tied to a single stage, it serves as a shared instrument across both the Data-Efficient and Experience-Driven stages, answering the same underlying question in each: whether the states the policy encounters are ones it has actually been trained on, which is central to a framework emphasizing robust real-world deployment over performance under idealized conditions.
 
\begin{figure}
    \centering
    \includegraphics[width=\linewidth]{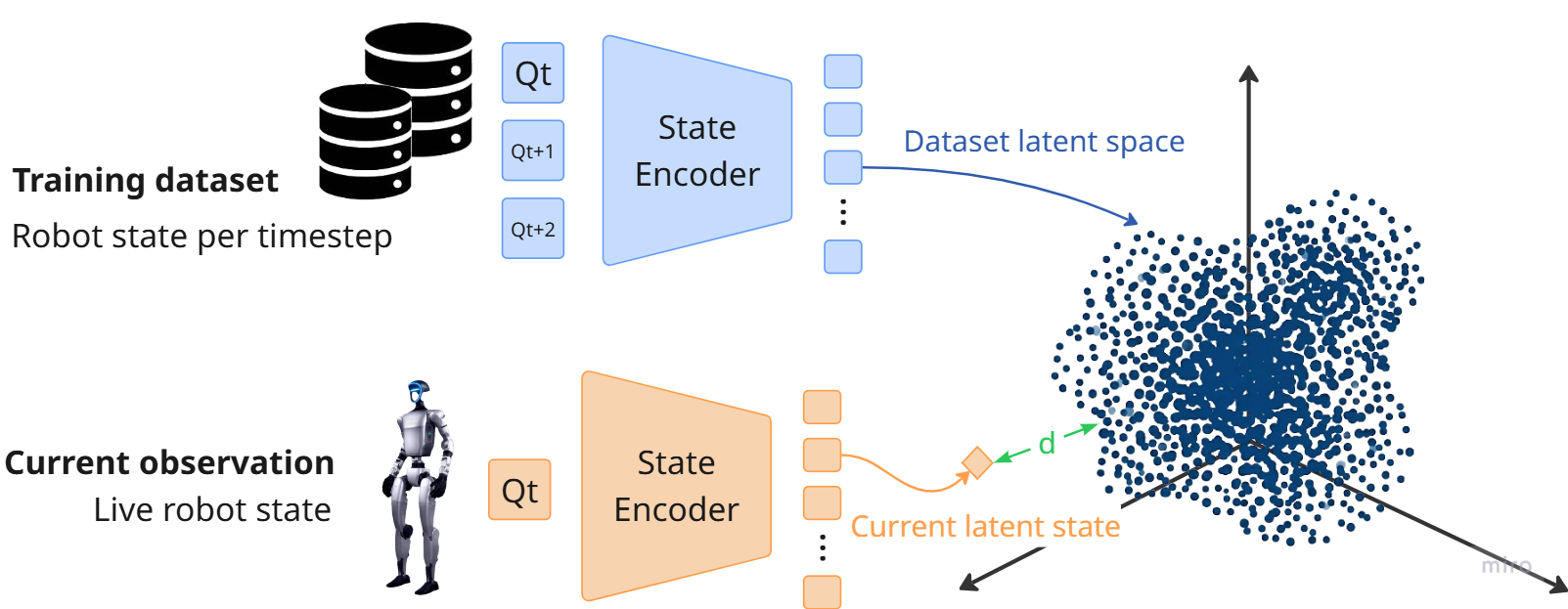}
    \caption{Overview of the state distribution analysis framework. Dataset states are projected into the VLA latent space using the state encoder and compared against the latent embedding of the current state observation.}
    \label{fig:ood}
\end{figure}

States are compared in the latent space of the fine-tuned GR00T state encoder, the representation the policy itself consumes when producing actions. Because policy learning shapes this encoder to emphasize control-relevant variation while compressing what can be ignored, distances here reflect task-relevant behavioral similarity rather than raw kinematic closeness: a state flagged as distant is one the policy is genuinely likely to mishandle, making the measure a predictor of behavioral reliability rather than an abstract geometric quantity. Figure~\ref{fig:ood} gives an overview.

Concretely, every timestep of the training dataset is passed through the state encoder $f_\phi(\cdot)$ to produce a latent representation $z = f_\phi(s)$, and a Gaussian Mixture Model (GMM) with $K$ components is fitted to the resulting embeddings $\{z_i\}_{i=1}^{N}$, approximating the empirical density as
\begin{equation}
  p(z) = \sum_{k=1}^{K} \pi_k \, \mathcal{N}(z \mid \mu_k, \Sigma_k),
\end{equation}
where $\pi_k$ are the mixture weights and $(\mu_k, \Sigma_k)$ each component's mean and covariance. The mixture formulation is important because a single task admits multiple valid configurations, so the demonstration manifold is inherently multi-modal; a unimodal fit would flag legitimate alternative strategies as anomalous. A new encoded state $z$ is scored by its Mahalanobis distance to the nearest mixture component,
\begin{equation}
  d(z) = \min_k \, (z - \mu_k)^\top \Sigma_k^{-1} (z - \mu_k),
\end{equation}
which accounts for each mode's covariance structure: deviations along high-variance directions, where the demonstrations already exhibit natural spread, contribute little, whereas deviations along low-variance directions are penalized strongly. This separates benign variation in joint configuration from genuine departures off the demonstrated manifold. The approach follows the latent-space density method for OOD detection introduced in \cite{lee2018simple}, here applied to a VLA state encoder's learned proprioceptive geometry rather than the features of a trained classifier.

Two additional outputs support interpretation. To make deviations attributable rather than merely detectable, per-joint z-scores and empirical percentile checks are computed alongside the aggregate distance, revealing which degrees of freedom drive a given deviation. For live visualization, PCA is fitted to the training embeddings and the first three principal components serve as a low-dimensional projection; the current state is projected under the same transformation and plotted against the training distribution, giving a direct geometric view of whether the robot lies inside the demonstrated region. Finally, thresholds separating in-distribution, soft-OOD, and hard-OOD regimes are derived from the empirical distribution of training distances rather than tuned by hand, allowing consistent classification across datasets and platforms.
 
Because it depends only on the state encoder and training embeddings, the tool is policy-agnostic, requires no modification to the VLA, and is reused unchanged across both DEED stages, in each case reporting shift relative to the current policy's training data.


\section{Experiments}
\label{sec:experiments}

\subsection{Experimental Setup}

Our experimental design targets real-world deployment rather than isolated laboratory metrics, emphasizing long-horizon operation: interaction with real objects, closed-loop control, recovery from failures, and repeated task execution.

\begin{figure}
    \centering
    \includegraphics[width=1\linewidth]{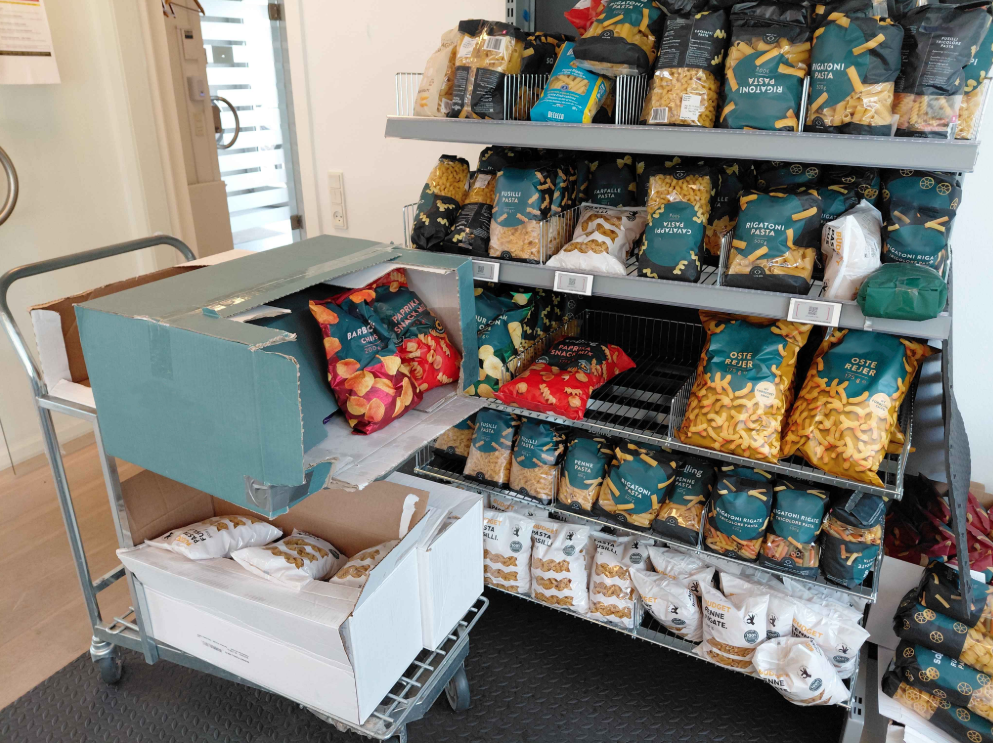}
    \caption{Deployment-oriented training setup for the restocking task. The model was trained using a realistic supermarket environment, including an actual shelf, trolley, and chip boxes, to expose the policy to the physical constraints and variability of the target deployment scenario while reducing the gap between training and real-world operation.}
    \label{fig:setup}
\end{figure}

We evaluate DEED on a real-world chip-restocking task using a Unitree G1-Edu humanoid with Dex-3 hands (Figure~\ref{fig:robot}). The robot must grasp a bag of chips from an opened product box on a trolley, rotate its torso toward an adjacent shelf, and place the bag, before rotating back to repeat the process. A supermarket store replica is arranged, containing a real shelf, trolley, and product box that keep the setup close to the target deployment scenario (Figure~\ref{fig:setup}). The task is performed primarily with the left arm while the right remains at rest, keeping both hands close to the torso and preserving balance under the Unitree locomotion controller.

\begin{figure*}[t]
    \centering
    \begin{subfigure}{0.245\linewidth}
        \includegraphics[width=\linewidth, trim=0 0 0 64px, clip]{"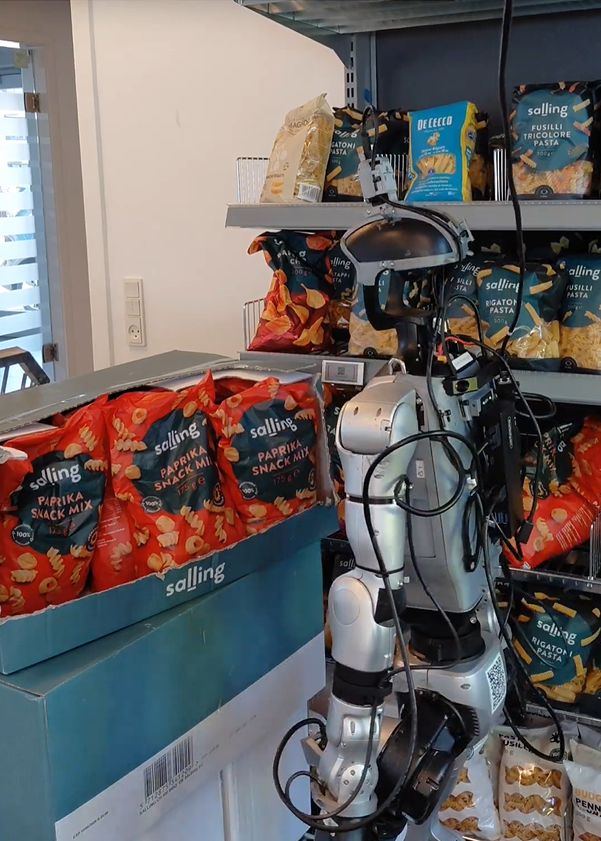"}
        \caption{Initial position}
    \end{subfigure}
    \hfill
    \begin{subfigure}{0.245\linewidth}
        \includegraphics[width=\linewidth, trim=0 0 0 64px, clip]{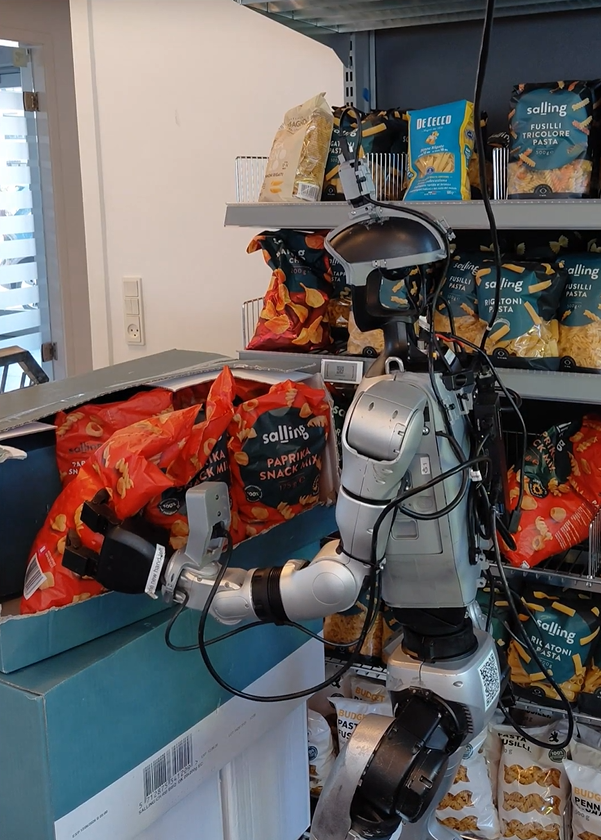}
        \caption{Picking}
    \end{subfigure}
    \hfill
    \begin{subfigure}{0.245\linewidth}
        \includegraphics[width=\linewidth, trim=0 0 0 64px, clip]{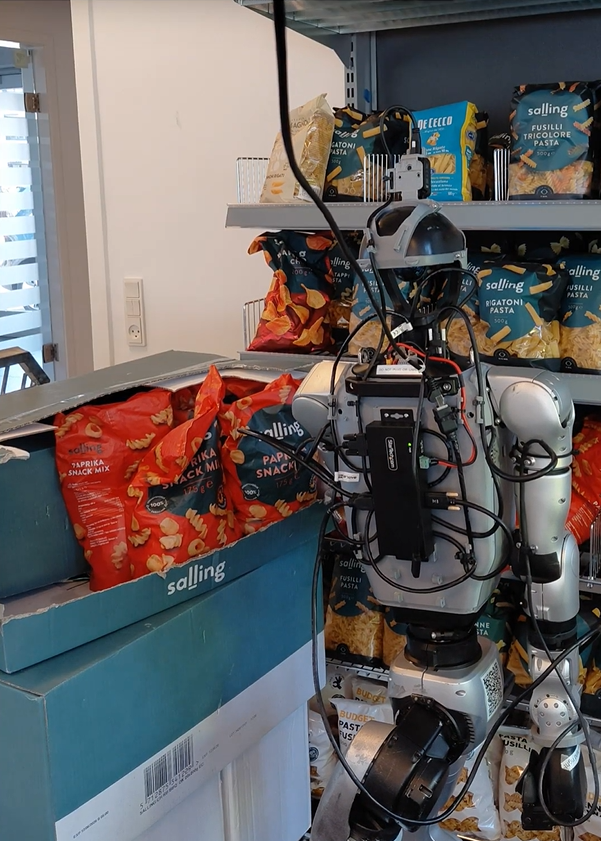}
        \caption{Placing}
    \end{subfigure}
    \hfill
    \begin{subfigure}{0.245\linewidth}
        \includegraphics[width=\linewidth, trim=0 0 0 64px, clip]{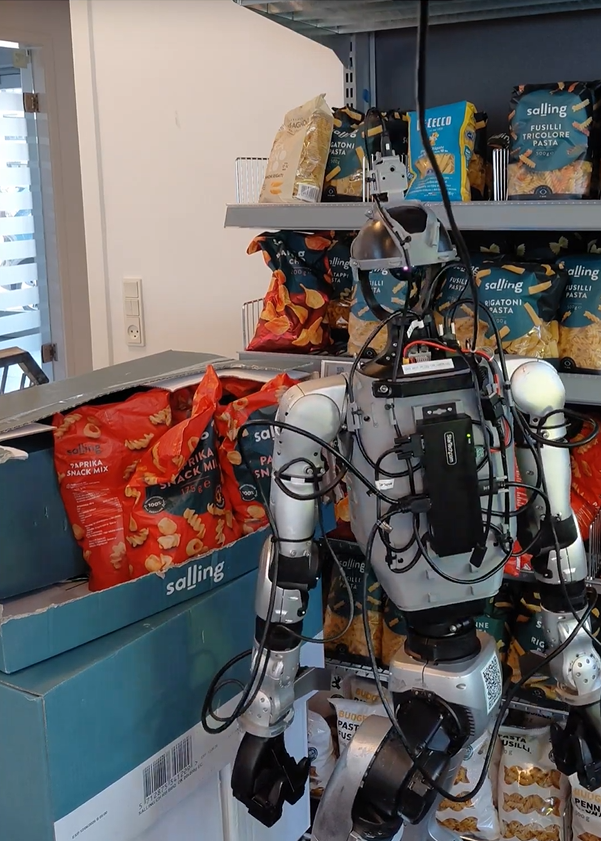}
        \caption{Rotating back}
    \end{subfigure}
    \caption{Representative frames from an evaluation episode of the chips restocking task. The robot grasps a bag of chips, transports it to the target shelf by rotating its torso, places it on the shelf, and subsequently returns to its initial configuration.}
    \label{fig:eval_episode_sequence}
\end{figure*}

\textbf{Embodiment and observations.}
Following the decoupled teleoperation setup, the action space is reduced from the native 32 dimensions to 20: seven joint commands per arm, a binary open/close command per hand, one waist joint, and three base velocity commands ($v_x$, $v_y$, $v_{\text{yaw}}$). The proprioceptive state is 17-dimensional, matching these components except the base velocities, which are passed to the locomotion policy but not observed. Visual input consists of three RGB streams, one head-mounted camera (Intel RealSense D435i) and two wrist-mounted cameras (RealSense D405), all at $640 \times 480$ and 30 FPS. Frequencies follow the hierarchy of Section~\ref{sec:DE_recipe}, with $f_t = 50$\,Hz, $f_{cam} = 30$\,Hz and $f_r = f_{ctrl} = 25$\,Hz.

\textbf{Collected Data.}
A total of 81 teleoperation demonstrations were collected, amounting to approximately 51.5 minutes of robot operation. All demonstrations use a single type of chips with no object variation, while the spatial distribution of the chip bags within the product box is kept balanced across episodes to ensure uniform coverage of the picking positions.

Data is collected through continuous episodes, where each episode may contain multiple picking and placing actions, ranging from restocking a single chip pack to restocking an entire box of six packs. Rather than segmenting full-box restocking into shorter clips, the whole sequence is retained as a single demonstration, allowing the policy to learn the transition between placing one bag and picking the next instead of repeatedly returning to a predefined starting configuration between actions.

In addition to the teleoperation data, 116 autonomous rollout episodes were gathered over the two RECAP iterations, comprising 41 successful executions and 75 failures for approximately 56.9 minutes of autonomous operation (27.4 minutes successful and 29.5 minutes failed). Overall, the dataset totals roughly 108.4 minutes of robot operation across teleoperation and autonomous rollouts.

\textbf{Training.}
All models are initialized from the publicly available \texttt{GR00T-N1.6-G1-PnPAppleToPlate} checkpoint. Model training, value function optimization, advantage calculation, and inference run on a single NVIDIA RTX 5090 workstation, which communicates with the robot over Ethernet during deployment.

The naive SFT baseline is fine-tuned on the same teleoperation demonstrations but with the native 32-dimensional action space, uncurated episodes, head-camera-only visual streams, no IA-VLA highlighting, and no inference-time smoothing.

\textbf{Evaluation.}
We report three metrics: task success rate, mean execution time over successful episodes, and the maximum number of consecutive bags restocked without a manual reset (in separate uninterrupted runs). Each evaluation episode is a single restocking attempt, successful if one bag is correctly grasped from the box and placed on the shelf without dropping or collision. Each condition is evaluated over 50 episodes starting from the same robot pose but with slight variation in the robot's position relative to the shelf and chips. We report the 95\% Wilson confidence intervals at this sample size and complement them with qualitative analysis.

\subsection{Results}

Table~\ref{tab:recap_results} reports the three metrics across the DEED pipeline stages. As a baseline we fine-tune \texttt{GR00T-N1.6} starting from the G1-specific public checkpoint \textit{GR00T-N1.6-G1-PnPAppleToPlate}. Despite the embodiment-specific initialization, this naive SFT is unable to perform the restocking task: without the DE engineering improvements it does not reliably grasp or place the bags, yielding a  0\% success rate.

Reaching a deployable policy required the combined design decisions of the Data-Efficient recipe, not fine-tuning alone: the reduced action space, added wrist cameras, matched control frequencies, curated demonstrations, and output smoothing together take the policy from non-functional to competent. The resulting \emph{DE policy} reaches 32\% (16/50) success at 24.30\,s per bag, is the first configuration to execute the full pick-rotate-place sequence autonomously, and is the base from which Experience-Driven refinement collects rollouts.
 
Building on this policy, a single Experience-Driven iteration improves both success rate and efficiency, raising success to 42\% (21/50) while reducing per-bag time to 22.37\,s. Qualitatively, the first RECAP iteration produces more direct approach trajectories and more reliable alignment during grasping and placement, consistent with the advantage-conditioned objective concentrating the policy around successful and human-corrected behaviors while teleoperation data still anchors the distribution.
 
A second iteration does not continue this trend: success drops to 22\% (11/50) despite a further small reduction in execution time (21.09\,s). We hypothesize this degradation stems from distributional effects during iterative refinement, discussed in Section~\ref{sec:discussion}. The in-distribution analysis supports this interpretation: the RECAP policies shift toward regions underrepresented in the teleoperation data despite remaining covered by the combined demonstration-plus-rollout dataset.
 
We note two important points on reliability. First, the 50-episode sample size limits statistical strength: the 95\% Wilson intervals for the base and first-iteration success rates (20.8-45.8\% and 29.4-55.8\%) overlap substantially, so the iteration-1 gain is indicative rather than conclusive, whereas the iteration-2 drop (12.8-35.2\%) shows less overlap. We therefore lean on the qualitative trends, which are consistent across runs. Second, across all configurations the policy fails to sustain multiple placements without a manual reset once refined (\emph{Max consecutive bags} of 1 for both RECAP iterations, versus 4 for the DE policy). This traces to the value function: the terminal completion is assigned a higher value than the intermediate poses of a reset, so reset transitions receive low advantage and are suppressed (Section~\ref{sec:discussion}).

The value-function design was selected via ablations over the VLM backbone (Eagle-3, Qwen2.5-VL, and PaliGemma), prompt specificity, state conditioning, and ensemble heads. Eagle-3 without state conditioning performed best, prompt specificity had negligible effect in the single-task setting, and ensemble heads exhibited near-zero disagreement (mean standard deviation: 0.009). Encoding the advantage as an additional state dimension significantly degraded performance, likely because it violated the zero-padded state convention established during pretraining.

\begin{table}[t]
\centering
\caption{DEED evaluation results on the chip restocking task. All policies start from \textit{GR00T-N1.6-G1-PnPAppleToPlate}; RECAP iterations add autonomous rollouts and human corrections to the teleoperation data.}
\label{tab:recap_results}
\setlength{\tabcolsep}{4pt}
\begin{tabularx}{\columnwidth}{@{}l c c c@{}}
\toprule
\textbf{Model} & \textbf{SR} & \textbf{Mean Duration (s)} & \textbf{Max bags} \\
\midrule
Naive SFT & 0\% (0/50)   & N/A   & 0 \\
DE policy                 & 32\% (16/50) & 24.30 & 4 \\
RECAP iteration 1         & 42\% (21/50) & 22.37 & 1 \\
RECAP iteration 2         & 22\% (11/50) & 21.09 & 1 \\
\bottomrule
\end{tabularx}
\end{table}

\subsection{Discussion}
\label{sec:discussion}

The two stages of DEED play complementary roles: the Data-Efficient stage turns an unusable checkpoint into a functioning policy from few demonstrations on a single GPU, while the Experience-Driven stage refines it from deployment experience, with one RECAP iteration yielding measurable if not statistically conclusive gains.
 
Unlike benchmarks with fixed poses and scene geometry, our evaluation involves unavoidable variability in robot initialization, object placement, and physical interaction, so success rates reflect end-to-end deployment robustness rather than performance under controlled assumptions. Within this context, the gains from experience-driven refinement do not compound across iterations: the second-iteration degradation is best explained as distributional drift, as self-generated rollouts come to dominate the training set, teleoperation demonstrations providing the broadest coverage of successful behavior lose relative influence, and advantage relabeling becomes increasingly tied to the policy's own sampling distribution, creating a feedback loop that progressively narrows the effective state-action distribution. Our in-distribution tool supports this, showing that the refined policies drift away from the demonstrated manifold even while remaining covered by the combined dataset.

How this degradation arises suggests where Experience-Driven refinement pays off. The pipeline rests on the assumption behind advantage-weighted offline learning: that similar states recur across episodes under consistent geometry, so advantage estimates are grounded in repeated observations rather than one-off configurations. This holds best in more static tasks with limited viewpoint variation, where the policy repeatedly encounters near-identical states and refinement can concentrate the distribution around successful behavior, yielding stable, compounding gains. In our restocking task, however, balancing, waist rotation, and changing foot placement make nominally identical configurations produce substantially different observations, so states recur less consistently and advantage estimation becomes noisier. More broadly, the benefit of refinement should diminish as a task demands more generalization: the greater the variation, the harder it is for self-generated rollouts to cover it, and the more readily refinement narrows onto a subset of behaviors. A practical implication is that maintaining a fixed mixture of teleoperation and rollout data, or periodically reintroducing demonstrations, is likely necessary to sustain gains as task variability grows.

A second, structural finding concerns reset behavior: refined policies tend to stop after a single placement rather than return to their start configuration, since the terminal completion state carries higher value than the intermediate poses of a reset, so reset transitions receive low or negative advantage and are suppressed during training. In practice we command the robot back to its starting pose rather than rely on the policy; addressing this within the learning framework would require changing the episode or reward definition, for instance by incorporating subtasks into the value function or explicitly assigning positive advantage to reset trajectories.
 
Finally, the in-distribution tool illustrates the value of treating monitoring as a first-class part of the framework. Because it measures shift in the policy's own latent representation and is agnostic to the specific VLA, it applies unchanged across both stages and across model versions. Here it served primarily as a diagnostic that turned an observed performance drop into a concrete, testable explanation, but the same signal could in principle be used online to detect when a deployed policy is leaving its training distribution and to trigger human intervention or data collection, closing the loop that DEED is built around.

\section{Conclusions}
\label{sec:conclusions}
We presented DEED, a systems-level framework for deploying VLA-based humanoids in real retail settings, demonstrated on a supermarket chip-restocking task with a Unitree G1-Edu and GR00T N1.6. Our main finding is that bridging the gap between pretrained VLAs and reliable deployment is primarily a systems integration challenge rather than an architectural one: without modifying the underlying model, careful data curation and targeted post-training turned a non-functional checkpoint into a capable policy on a single GPU, with data quality mattering more than data volume.

The two stages play complementary roles. The Data-Efficient stage does the bulk of the work (0\% to 32\% success), while a single Experience-Driven iteration raises success to 42\%. Gains did not compound: a second iteration reduced robustness, likely from distributional drift as self-generated rollouts came to dominate the data, underscoring the need to maintain a balanced mixture of demonstrations and rollouts during continual learning, especially for high-variability humanoid tasks. Our in-distribution analysis provides a model-agnostic signal for diagnosing such degradation and could be extended to trigger intervention or data collection online.

These conclusions are limited by a single task, one platform, a binary success metric, few evaluation episodes, and two refinement iterations. Future work should evaluate DEED across more tasks and platforms, develop continual-learning strategies that preserve the demonstration-rollout balance, and couple deployment monitoring with automatic adaptation.


\bibliographystyle{IEEEtran}
\bibliography{refs}

\end{document}